\pdfoutput=1

\documentclass[11pt]{article}

\usepackage[final]{acl}

\usepackage{times}
\usepackage{latexsym}

\usepackage[T1]{fontenc}

\usepackage[utf8]{inputenc}

\usepackage{microtype}

\usepackage{inconsolata}

\usepackage{amsmath} 
\DeclareMathOperator{\inv}{inv}

\usepackage[utf8]{inputenc}

\usepackage{graphicx}
\usepackage{soul}
\usepackage{xcolor}
\definecolor{gael}{RGB}{220, 0, 255}

\newcommand{\hlpink}[1]{\sethlcolor{pink}\hl{#1}}  
\newcommand{\hlgreen}[1]{\sethlcolor{green}\hl{#1}}   

\usepackage{xspace} 
\usepackage{multirow}
\usepackage{diagbox}
\usepackage{algorithm}
\usepackage{algpseudocode}

\newcommand{\app}[0]{\mbox{\textit{QUARTZ}}\xspace}

\usepackage{booktabs}
\usepackage{tabularray}

\title{
\app: QA-based Unsupervised Abstractive Refinement for Task-oriented Dialogue Summarization}

\author{
 \textbf{Mohamed Imed Eddine Ghebriout\textsuperscript{1}}\thanks{Corresponding author.},
 \textbf{Gaël Guibon\textsuperscript{1,2}},
 \textbf{Ivan Lerner\textsuperscript{3,4,5}},
 \textbf{Emmanuel Vincent\textsuperscript{1}}
\\
\\
 \textsuperscript{1}\small Université de Lorraine, CNRS, Inria, LORIA, F-54000 Nancy, France\\ 
 \textsuperscript{2}\small Université Sorbonne Paris Nord, CNRS, Laboratoire d’Informatique de Paris Nord, LIPN, F-93430 Villetaneuse, France\\
 \textsuperscript{3}\small Inserm, Centre de Recherche des Cordeliers, Université Paris Cité, Sorbonne Université, F-75006 Paris, France\\
 \textsuperscript{4}\small HeKA, Inria Paris, F-75012 Paris, France\\
  \textsuperscript{5}\small Department of Medical Informatics, Assistance Publique Hôpitaux de Paris, Georges Pompidou European Hospital
  \\
 \small{
   \{imed-eddine.ghebriout, gael.guibon\}@loria.fr \qquad \quad \{ivan.lerner, emmanuel.vincent\}@inria.fr
 }
}

\begin{document}
\maketitle
\begin{abstract}
Dialogue summarization aims to distill the core meaning of a conversation into a concise text. This is crucial for reducing the complexity and noise inherent in dialogue-heavy applications. 
While recent approaches typically train language models to mimic human-written summaries, such supervision is costly and often results in outputs that lack task-specific focus limiting their effectiveness in downstream applications, such as medical tasks.
In this paper, we propose \app, a framework for task-oriented utility-based dialogue summarization. \app starts by generating multiple summaries and task-oriented question-answer pairs from a dialogue in a zero-shot manner using a pool of large language models (LLMs). The quality of the generated summaries is evaluated by having LLMs answer task-related questions before \textit{(i)} selecting the best candidate answers and \textit{(ii)} identifying the most informative summary based on these answers. Finally, we fine-tune the best LLM on the selected summaries. When validated on multiple datasets, \app demonstrates its effectiveness by achieving competitive results in various zero-shot settings, rivaling fully-supervised State-of-the-Art (SotA) methods. \textit{Code and supplementary material are publicly available}\footnote{https://github.com/Mohamed-Imed-Eddine/QUARTZ}.
\end{abstract}

\section{Introduction}
Automatic Text Summarization (ATS) is the task of generating a concise summary from a lengthy text. It can be \textit{(extractive)} -- selecting and concatenating key sentences -- or \textit{(abstractive)} -- generating new condensed formulation that resemble human-written summaries \citep{lin2019abstractive}. 
While early ATS methods relied on graph-based \citep{mihalcea2004textrank} and frequency-based techniques \citep{alsaedi2016temporal}, the emergence of large language models (LLMs) has transformed the field, achieving breakthroughs across domains, including healthcare \citep{van2024adapted}.
Dialogue summarization, a subfield of ATS, targets key information distillation from conversations. It has gained prominence due to its applicability in real-world settings such as customer service \citep{feigenblat2021tweetsumm}, business meetings \citep{rennard2023abstractive}, and healthcare \citep{abacha2023empirical}, where the input is often less structured than typical written texts like news articles. Summarizing such dialogues helps simplify complex interactions and supports downstream tasks like decision-making and automation.
Dialogues introduces challenges absent in standard ATS: disfluencies, speaker turns, redundancy, verbosity, and loosely structured content. These characteristics make traditional ATS methods, designed for more structured text, less effective, often resulting in low-quality summaries \citep{zechner2002automatic,fengsurvey}. 
Recently, the capability of LLMs to perform dialogue summarization has been assessed by framing it as a sequence-to-sequence task \citep{van2024adapted, tian2024dialogue}. Fine-tuning LLMs for task-oriented summarization has demonstrated promising results, even surpassing human performance \citep{van2024adapted}. However, this approach has limited practicality due to the reliance on costly human-written summaries. Moreover, while the vast amount of text seen during training enables LLMs to be powerful \textit{“multitask learners”}, their lack of robust mechanisms for maintaining topic focus and factual consistency makes them prone to topic-deviations, often resulting in incoherent summaries \citep{tonmoy2024comprehensive}.

In this paper we present \app, a framework for unsupervised, task-oriented abstractive dialogue summarization. Its task-oriented design enables better control over LLM outputs by maintaining topic focus and factual consistency, while its unsupervised nature eliminates the need for reference summaries or domain-specific knowledge beyond that used to design task-specific LLM prompts. Our framework encompasses the following novelties and contributions:

\begin{enumerate}
    \item To the best of our knowledge, \app is the first framework to enhance LLM-based dialogue summarization in a reference-free setting.
    \item \app generates multiple candidate summaries per dialogue and introduces a new two-stage selection process to identify the best summary given a set of task-oriented questions and LLM-derived answers from the summaries. In stage one, it identifies the best answers for each summary-question pair. In stage two, it selects the most informative summary based on these answers.
    \item Even without any gold summary, fine-tuning on the selected generated summaries further boosts summarization effectiveness.
    \item Our framework is applicable to real-world scenarios, for instance to provide assistance and facilitate the summarization of clinical conversations.

\end{enumerate}

\begin{figure*}[ht!]
\centering
    \includegraphics[width=0.99\textwidth]{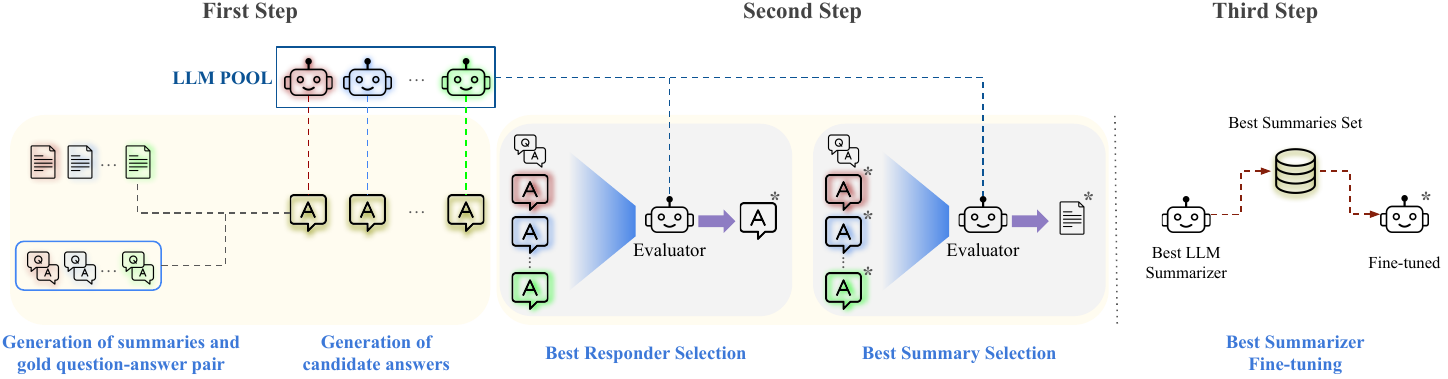}
   \caption{An overview of \app for unsupervised task-oriented dialogue summarization.} 
    \label{fig:framework}
\end{figure*}

\section{Related Work}
\subsection{Dialogue Summarization}
\label{sec:relatedwork_dialogsum}
Dialogue summarization has attracted growing interest due to its utility in domains such as healthcare and business. Unlike document summarization, it involves additional challenges like frequent semantic shifts, redundancies and multi-speaker dynamics. 
To address these, \citet{zhao2021todsum} leveraged dialogue state tracking to enhance coherence, while \citet{tian2024dialogue} used a Mixture-of-Experts (MoE) \cite{jacobs1991adaptive} LLM to process utterances via specialized experts and fusing their representations to produce summaries. Despite these efforts, methods still struggle with coherence, information coverage and overall grasp of dialogue structure \citep{zhao2021todsum}. Task-oriented dialogue summarization offers a more focused perspective. For instance, \citet{wang2023instructive} proposed instruction-guided summarization by deriving targeted queries from reference summaries to steer LLM generation. However most task-oriented approaches require a decent amount of dialogue data with human-written summaries to achieve cutting-edge performance \citep{zou2021low}, low-resource settings have been addressed via data augmentation based on pre-trained language models. \citet{ouyang2023compositional} generated dialogue-summary pairs to create new training instances, and \citet{liu2022data} replaced sections from the input dialogues and summaries using generated text. In contrast, the area of unsupervised dialogue summarization, which we tackle in this paper, remains largely unexplored.

\subsection{LLMs for Data Augmentation}
Data Augmentation (DA) refers to strategies aimed at enhancing training data diversity without additional manual collection \citep{feng2021survey}. It has been widely applied in Natural Language Processing (NLP) tasks, including low-resource machine translation \citep{xia2019generalized}, summary grounded conversation generation \citep{gunasekara2021summary}, and Question Answering (QA) \citep{guo2023improving}. Concerning the latter, \citet{yang2019data} proposed a DA method to enhance BERT fine-tuning for open-domain QA using distant supervision, and \citet{riabi2021synthetic} employed a question generation model to augment examples for cross-lingual QA. The rise of LLMs has revolutionized DA, enabling the generation of high-quality synthetic data and reducing the data collection and labeling costs \citep{tang2023does}. Examples include GPT-3 generating synthetic medical dialogue summaries for models training \citep{chintagunta2021medically}, and GPT-4 producing an instruction-following dataset for LLaMA fine-tuning \citep{peng2023instruction}.

\subsection{QA Evaluation using LLMs}
QA evaluation has emerged as an effective proxy for assessing factuality in summaries \citep{wang2024evaluating}.  
It ensures that key information in the original text remains accessible and accurate in the summary by comparing candidate answers to predefined gold answers. 
Historically, lexical matching dominated QA evaluation through techniques like Exact-Match \citep{izacard2021leveraging} or N-gram matching \citep{chen2017reading}. 
Subsequently, similarity metrics such as BERT Matching \citep{bulian2022tomayto} emerged to assess how closely a candidate answer aligns with the core meaning of the gold answer. 
More recently, LLMs have proven highly effective for this task, supported by studies demonstrating performance on par with human evaluators \citep{bavaresco2024llms,tornberg2023chatgpt, song2024finesure}, offering a cost-effective and scalable alternative. 
This is typically achieved by conditioning the LLM’s behavior using a system prompt, the question, the gold answer and the candidate answer to be assessed \citep{wang2024evaluating, kamalloo2024towards}. 
This process can be further enhanced through prompt engineering techniques such as In-Context Learning \citep{brown2020language} and Chain-of-Thoughts \citep{chainofthought}. 
However, long contexts, can hinder LLM focus, leading to unreliable assessments. This phenomenon was inspected by \citet{liu2024lost}, who showed that LLMs process different parts of a long context inconsistently, and often neglect information in the middle. To solve this, repeated prompting and score aggregation tend to be a remedy \citep{tang2024found}.

\section{Our Approach}
The overall workflow of our approach is illustrated in Figure~\ref{fig:framework}.  Inspired by recent findings in summarization, showing that LLM summaries are significantly preferred by human evaluators \citep{pu2023summarization}, \app starts by \textbf{(First Step)} prompting a pool of LLMs -- rather than a single model -- helps mitigate model-specific tendencies and enables the generation of varied task-oriented summaries along with gold question-answer pairs from the input dialogue. Tailored prompts (Appendix~\ref{sec:appendix_prompts}) are used to ensure focus on task-relevant information. We also utilize LLMs as responders to answer the generated questions using the summaries, rather than using the original dialogues. We then \textbf{(Second Step)} judge the quality of the generated summaries by assessing the candidate answers using a two-stage evaluation process. In the first stage, 
 LLMs are employed as rankers to select the best candidate answer for each question and summary. In the second stage, 
 LLM ranking reveals the best summary for each dialogue. Finally, \textbf{(Third Step)} the best LLM is fine-tuned on the selected summaries to further improve its task-oriented summarization performance. We detail the different stages of \app and its evaluation in the following.

\subsection{First Step: Generation}
\paragraph{Generation of Task-Oriented Summaries.}
\leavevmode \\
A task-oriented dialogue is framed as a multi-turn conversation between two or more participants aimed at achieving a specific goal.
We prompt each model $l_\text{S}$ from the pool $L$ to generate one summary $S_{i,l_\text{S}}$ for each dialogue $D_i$, conditioning this process on a carefully crafted task prompt $T$ that fetches the task-specific information.

\paragraph{Generation of Gold Question-Answer Pairs.}
Similarly, each model $l_\text{Q} \in L$ is prompted to generate a set of gold question-answer (QA) pairs from each dialogue $D_i$, focusing on questions that ought to be answerable using the task-oriented dialogue summary. The gold QA pairs generated by all LLMs are merged into a single set of $J_i$ gold QA pairs $(Q_{i,j},A_{i,j})$ indexed by $j$.

\paragraph{Generation of Candidate Answers.}
\leavevmode\newline 
With summaries and questions in place, we next derive candidate answers from the generated summaries. Indeed, a factual summary should retain the key information needed to answers all task-relevant questions.
To do so, we prompt each model $l_\text{R}\in L$ to generate one candidate answer $\hat{A}_{i,j,l_\text{S},l_\text{R}}$ for each question $Q_{i,j}$ from each summary $S_{i,l_\text{S}}$. Recognizing the self-bias phenomenon -- where LLMs may easily respond to its own generated question \citep{xu2024pride} -- , and aware of the risk of amplifying this bias, we use all LLMs in the pool as responders, including the question generator LLM itself. This reduces bias and provides a fair way of answering the questions. 

\subsection{Second Step: Two-Stage Evaluation}

\paragraph{First-Stage Evaluation \textit{(Best Responder Selection).}}
\label{first_stage}
At this stage we identify the best candidate answers for each question-summary pair. To do so, we use each model $l_\text{E}\in L$ as an evaluator, receiving the question $Q_{i,j}$, the gold answer $A_{i,j}$, and the set of $|L|$ candidate answers $\{\hat{A}_{i,j,l_\text{S},l_\text{R}}\}_{l_\text{R}\in L}$ obtained from summary $S_{i,l_\text{S}}$, and it is instructed to provide a ranking (defined as a permutation $\sigma:\{1, \dots, |L|\} \mapsto \{1, \dots, |L|\}$) for the candidate answers based on their relevance and correctness. To overcome the self-preference bias \citep{panickssery2024llm} by which an LLM evaluator scores its own outputs higher than others, we use all LLMs in the pool as evaluators, including the LLM responder itself. We do this by prompting the LLMs multiple times for ranking while randomly altering the order of the candidate answers and subsequently aggregating the output ranks. This ensures that no LLM has a systematic advantage over the others, and has been theoretically and empirically proven by \citet{tang2024found} to converge to the true ranking. Therefore, we prompt each LLM $N$ times to obtain $N$ estimated rankings $\hat{\sigma}_{i,j,l_\text{S},l_\text{E},n}$ indexed by $n$. 
The optimal ranking $\sigma_{i,j,l_\text{S},l_\text{E}}$ is the one whose sum of Kendall tau distances \citep{kendall1938new} to all estimated rankings is minimum \citep{tang2024found}:
 \begin{equation}
      \sigma_{i,j,l_\text{S},l_\text{E}} := \arg\min_{\sigma}\sum_{n=1}^{N} d_\kappa(\hat{\sigma}_{i,j,l_\text{S},l_\text{E},n}, \sigma).
 \end{equation}
The Kendall tau distance $d_\kappa(.,.)$ quantifies the dissimilarity between two rankings, specifically representing the number of \textit{discordant} pairs:
\begin{equation}
    d_{\kappa}(\sigma_1, \sigma_2):= \sum_{k=1}^{|L|} \inv(\sigma_1^{-1} \circ \sigma_2)_k
    \label{eq:kendal_distance}
\end{equation}
where

\begin{equation}
    \inv(\sigma)_k := \#\{k':\sigma(k') > \sigma(k), k'<k\}.
    \label{eq:inversion_vector}
\end{equation}

After obtaining the optimal ranking $\sigma_{i,j,l_\text{S},l_\text{E}}$ for each question $j$, we compute the Mean Reciprocal Rank (MRR) \citep{radev-etal-2002-evaluating} of each LLM responder $l_\text{R}$ over all questions. The MRR is calculated as follows: 
\begin{equation}
\text{MRR}_{i,l_\text{S},l_\text{R},l_\text{E}}  =  \dfrac{1}{J_i} \sum_{j=1}^{J_i} \dfrac{1}{\sigma^{-1}_{i,j,l_\text{S},l_\text{E}}(l_\text{R})}.
\label{eq:mrr_formula_R_E}
\end{equation}
Note that the MRR still depends on the evaluator $l_\text{E}$. To select the best responder independently of the evaluator, we compute the total score
\begin{equation}
\text{Score}_{i,l_\text{S},l_\text{R}} = \sum_{l_\text{E}\in L}
\alpha_{l_\text{R},l_\text{E}}\,\text{MRR}_{i,l_\text{S},l_\text{R},l_\text{E}}
\label{eq:score_ls_lr}
\end{equation}
where the weighting factor 
\begin{equation}
\alpha_{l_\text{R},l_\text{E}} = 
\begin{cases}
0.8 & \text{if } l_\text{R}=l_\text{E} \\
1 & \text{otherwise}
\end{cases}
\label{eq:alpha_lr_le}
\end{equation}
penalizes when the same model serves as both responder and evaluator. Finally, for each dialogue $D_i$ and summary $S_{i,l_\text{S}}$, we select the responder with the highest score and use the corresponding candidate answers $\hat{A}_{i,j,l_\text{S}}^*$ as inputs to the second stage.

\paragraph{Second-Stage Evaluation \textit{(Best Summary Selection).}}
After selecting the best candidate answers for each summary, we determine the best summary for each dialogue. Similarly to above, we use each model $l_\text{E}\in L$ as an evaluator. The evaluator is provided with a question $Q_{i,j}$, its gold answer $A_{i,j}$, and the set of $|L|$ best candidate answers $\{\hat{A}_{i,j,l_\text{S}}^*\}_{l_\text{S}\in L}$ obtained from all summaries, and is instructed to rank them. We repeat this $N$ times and derive the final ranking $\sigma_{i,j,l_\text{E}}$ as above. The MRR of each LLM summarizer $l_\text{S}$ is then computed over all questions as
\begin{equation}
\text{MRR}_{i,l_\text{S},l_\text{E}}  =  \dfrac{1}{J_i}  \sum_{j=1}^{J_i} \dfrac{1}{\sigma^{-1}_{i,j,l_\text{E}}(l_\text{S})}.
\label{eq:mrr_formula_S_E}
\end{equation}
To select the best summary independently of the evaluator, we compute the total score
\begin{equation}
\text{Score}_{i,l_\text{S}} = \sum_{l_\text{E}\in L} \alpha_{l_\text{S},l_\text{E}}\,\text{MRR}_{i,l_\text{S},l_\text{E}}
\end{equation}
with a similar weighting factor as in Equation~\eqref{eq:alpha_lr_le}.
Finally, for each dialogue $D_i$, we select the summary $S_i^*$ with the highest score.
 
\begin{table}[!b]
    \centering
    \scalebox{0.8}{
    \begin{tabular}{lcccc}
    
        \toprule
        \textbf{Dataset} & \textbf{} & \textbf{\# Dial.} & \textbf{Avg. Len.} & \textbf{Avg. Turns} \\
        \toprule
        \multirow{3}{*}{\bf SAMSum} & Train & 14,732 & 93.8 & 11.2 \\ 
                                & Valid & 818 & 91.6 & 10.8 \\ 
                                & Test & 819 & 95.5 & 11.3 \\ \cmidrule{2-5}

        \multirow{3}{*}{\bf DialogSum} & Train & 12,460 & 131.0 & 9.5  \\ 
                                & Valid & 500 & 129.3 & 9.4 \\ 
                                & Test & 1,500 & 134.5 & 9.7 \\ \cmidrule{2-5}

        \multirow{3}{*}{\bf MTS-Dialog} & Train & 1,201 & 87.99 & 8.9  \\ 
                                & Valid & 100 & 77.46 & 7.7 \\ 
                                & Test & 200 & 87.69 & 9.1 \\ \cmidrule{2-5}
                                    
        {\bf SimSAMU} & {-} & {61} & {502.47} & {50.63} \\ \bottomrule
    \end{tabular}
    }
    \caption{Dataset statistics. \textbf{``\# Dial.''} refers to the number of dialogues, \textbf{``Avg. Len.''} represents the average number of tokens per dialogue, and \textbf{``Avg. Turns''} indicates the average number of turns per dialogue.}
    \label{tab:datasets}
\end{table}

\subsection{Third Step: Fine-Tuning}
\label{sec:finetune}
Eventually, we fine-tune the LLM summarizer $l_\text{S}^*$ that produced most selected summaries. Although this step involves learning, it remains unsupervised, hence no reference summaries or human labels are used. Training pairs consist of input dialogues and their best selected summaries:
\begin{equation}
  \max_{\theta} \sum_{i=1}^I \log P(S_i^* \mid D_i, T, \theta)
  \label{eq:optim_summ}
\end{equation}
where $\theta$ are model parameters and $T$ is the task-specific prompt.

\section{Experimental Settings}
In the following, we explain our experimental evaluation protocol, including the datasets we used and the implementation details.

\subsection{Datasets}
We conduct experiments on four task-oriented datasets, summarized in Table~\ref{tab:datasets}:

\noindent{\large \textcircled{\small 1}} SAMSum \citep{gliwa2019samsum} contains messenger-style conversation between two or more people with their corresponding summaries created by linguists to simulate real-life and task-specific scenarios.

\noindent{\large \textcircled{\small 2}} DialogSum \cite{chen2021dialogsum} includes real-life dialogues spanning diverse task-oriented scenarios (e.g., business negotiation and doctor visits), with a perspective to support downstream applications for both business and personal use.

\noindent{\large \textcircled{\small 3}} MTS-Dialog \citep{abacha2023empirical} features doctor-patient dialogues paired with real-world clinical notes covering multiple specialties (e.g., Neurology, Immunology). Each dialogue is accompanied by a header (e.g., diagnosis, exam) that guides medical report generation.

\noindent{\large \textcircled{\small 4}} SimSAMU \citep{NUN2025108857} is a French medical dispatch dialogue dataset comprising the transcripts of 3 hours of audio recordings of simulated emergency dispatch dialogues across various incidents. Given the characteristics of this dataset displayed in Table \ref{tab:datasets} (e.g., lengthy dialogues with multiple turns), 
task-oriented summaries help assess incident severity and support timely, accurate triage decisions.

\subsection{Implementation Details}
We utilize a pool of three LLMs: \textit{Llama-3.1-8B-Instruct} \citep{dubey2024llama}, \textit{Gemma-2-9b-it} \citep{team2024gemma}, and \textit{Qwen2-7B-Instruct} \citep{yang2024qwen2}. Each model generates 10 to 15 gold QA pairs, which are merged into a unified set of $J_i$ pairs. For LLM ranking, we found that rank consistency is satisfactory for $N=5$. Further implementation details can be found in Appendix~\ref{sec:appendix_implementation_details}.

\begin{table*}[ht]
    \centering
    \resizebox{\textwidth}{!}{
    \begin{tabular}{lccccc}
        \textbf{Method} & \textbf{R-1} $\uparrow$ & \textbf{R-2} $\uparrow$ & \textbf{R-L} $\uparrow$ & 
        \textbf{BLEU} $\uparrow$ & \textbf{BERT-Score} $\uparrow$ \\ \midrule
        \multicolumn{6}{c}{\bf Dataset {\large \textcircled{\small 1}} SAMSum -- Chit-Chat Domain (Casual, open-domain conversations)} \\ \hline \toprule
        \multicolumn{6}{c}{Supervised Dialogue Summarization} \\ \midrule
         {MoE \citep{tian2024dialogue}} & \textbf{55.93} & 30.86 & \textbf{52.02} & \textbf{26.03} & \textbf{75.66} \\ 
         {InstructDS \citep{wang2023instructive}} & 55.30 & \textbf{31.30} & 46.70 & - & 55.50 \\ 
         {MoCa \citep{zhang2022momentum}} & 55.13 & 30.57 & 50.88 & - & - \\ \midrule
        \multicolumn{6}{c}{Unsupervised Dialogue Summarization} \\ \midrule 
         {Llama-3.1-8B-Instruct \citep{dubey2024llama}} & 29.93\small{$\pm 0.98$} & 10.89\small{$\pm 0.93$} & 22.38\small{$\pm 0.78$} & 4.85\small{$\pm 0.33$} & 58.19\small{$\pm 0.51$} \\
        {DeepSeek-R1-Distill-Llama-8B \citep{guo2025deepseek}} & 31.69\small{$\pm 1.20$} & 10.70\small{$\pm 0.93$} & 23.91\small{$\pm 0.96$} & 3.45\small{$\pm 0.37$} & 59.16\small{$\pm 0.56$} \\
        {DeepSeek-R1-Distill-Qwen-14B \citep{guo2025deepseek}} & 36.04\small{$\pm 1.14$} & 12.79\small{$\pm 0.98$} & 27.68\small{$\pm 0.99$} & 5.56\small{$\pm 0.51$} & 61.67\small{$\pm 0.56$} \\  
        {AAF \citep{li2025adaptive}} & 44.57 & 19.04 & 35.65 & - & - \\  \hline   
         {\app (Best summarizer: Llama 3.1)} & \textbf{61.37}\small{$\pm 1.36$} & \textbf{38.54}\small{$\pm 1.75$} & \textbf{53.12}\small{$\pm 1.53$} & \textbf{31.07}\small{$\pm 0.89$} & \textbf{77.95}\small{$\pm 0.63$} \\[2mm]

        \multicolumn{6}{c}{\bf Dataset {\large \textcircled{\small 2}} DialogSum -- Task-Oriented Domain (Real-life dialogues across various scenarios)} \\ \hline  \toprule
         \multicolumn{6}{c}{Supervised Dialogue Summarization} \\ \midrule
         {MoE \citep{tian2024dialogue}} & \textbf{49.82} & \textbf{24.80} & \textbf{47.34} & \textbf{18.41} & \textbf{68.48} \\
         {InstructDS \citep{wang2023instructive}} & 47.8 & 22.2 & 39.4 & - & 47.0 \\ \midrule
         \multicolumn{6}{c}{Unsupervised Dialogue Summarization} \\ \midrule
         {Llama-3.1-8B-Instruct \citep{dubey2024llama}} & 21.35\small{$\pm 0.79$} & 7.36\small{$\pm 0.41$} & 16.46\small{$\pm 0.60$} & 3.09\small{$\pm 0.19$} & 53.70\small{$\pm 0.39$} \\			
        {DeepSeek-R1-Distill-Llama-8B \citep{guo2025deepseek}} & 27.11\small{$\pm 0.77$} & 7.53\small{$\pm 0.74$} & 20.37\small{$\pm 1.09$} & 2.79\small{$\pm 0.21$} & 54.27\small{$\pm 0.40$} \\
        {DeepSeek-R1-Distill-Qwen-14B \citep{guo2025deepseek}} & 29.03\small{$\pm 0.77$} & 8.72\small{$\pm 0.62$} & 21.99\small{$\pm 0.87$} & 3.44\small{$\pm 0.32$} & 55.47\small{$\pm 0.43$} \\
        {AAF \citep{li2025adaptive}} & \textbf{41.38} & 15.09 & 32.52 & - & - \\  \hline 
         {\app (Best summarizer: Llama 3.1)} & 39.73\small{$\pm 1.08$} & \textbf{16.02}\small{$\pm 0.87$} & \textbf{32.68}\small{$\pm 0.92$} & \textbf{15.73}\small{$\pm 0.65$} & \textbf{68.72}\small{$\pm 0.44$} \\[2mm]

        \multicolumn{6}{c}{\bf Dataset {\large \textcircled{\small 3}} MTS-Dialog -- Medical Domain (Doctor-patient interactions with clinical notes)} \\ \hline \toprule 
        \multicolumn{6}{c}{Supervised Dialogue Summarization} \\ \midrule
         {BART-GS-DA \citep{abacha2023empirical}} & \textbf{42.52} & \textbf{17.50} & \textbf{34.90} & - & \textbf{40.80} \\ \midrule

        \multicolumn{6}{c}{Unsupervised Dialogue Summarization} \\ \midrule
        {GPT3-ICL \citep{suri2023healthmavericks}} & 19.87 & 8.67 & 15.60 & - &57.03  \\ 
         {Llama-3.1-8B-Instruct \citep{dubey2024llama}} & 28.29\small{$\pm 2.37$} & 10.19\small{$\pm 1.53$} & 21.14\small{$\pm 1.79$} & 7.57\small{$\pm 1.28$} & 55.82\small{$\pm 1.54$} \\ 
        {DeepSeek-R1-Distill-Llama-8B \citep{guo2025deepseek}} & 17.76\small{$\pm 2.31$} & 5.68\small{$\pm 1.14$} & 13.02\small{$\pm 1.71$} & 2.46\small{$\pm 0.60$} & 47.54\small{$\pm 1.49$} \\
        {DeepSeek-R1-Distill-Qwen-14B \citep{guo2025deepseek}} & 26.31\small{$\pm 2.85$} & 10.17\small{$\pm 1.84$} & 19.78\small{$\pm 2.33$} & 4.29\small{$\pm 1.01$} & 52.84\small{$\pm 1.69$}  \\ \hline
         {\app (Best summarizer: Llama 3.1)} & \textbf{34.63}\small{$\pm 2.88$} & \textbf{13.98}\small{$\pm 2.19$}  & \textbf{27.72}\small{$\pm 2.60$} & \textbf{12.75}\small{$\pm 2.12$} & \textbf{61.18}\small{$\pm 1.73$}\\

    \end{tabular}
    }
    \caption{Performance evaluation of \app on the test sets of SAMSum, MTS-Dialog, and DialogSum.}
    \label{tab:all_results}
\end{table*}

\subsection{Evaluation Protocol}
We evaluate the test summaries using ROUGE (R-1, R-2 and R-L) \cite{lin-2004-rouge}, BLEU \cite{papineni2002bleu} and BERT-Score \citep{bert-score}. Jackknife resampling \citep{efron1981jackknife} is used to provide 95\% confidence intervals. Evaluation also includes human judgments (coherence, consistency, fluency, relevance) and LLM-as-Judge scoring.\\

\noindent \textbf{Baselines.} As unsupervised SotA approaches for task oriented dialogue summarization are scarce (Section \ref{sec:relatedwork_dialogsum}), we benchmark our framework mostly against fully supervised methods. 
\citet{tian2024dialogue} proposed a role-oriented routing based MoE summarizer. \citet{wang2023instructive} proposed an instruction-tuning method for instructive dialogue summarization, whereas \citet{zhang2022momentum} introduced momentum calibration to better align the model generation with reference summaries. \citet{abacha2023empirical} leveraged augmented pre-training followed by fine-tuning on supervised data. In contrast, GPT3-ICL \citep{suri2023healthmavericks} prompted OpenAI GPT3 with similar conversations and summaries for each test sample. While in \citet{li2025adaptive}, adaptive augmentation fusion is performed throughout training epochs of summarization model on summaries generated by a DA model, guided by a subset of the reference summaries. We also evaluate DeepSeek-R1 distilled models \citep{guo2025deepseek}, available in 8B and 14B sizes, to investigate the impact of the Chain of Thought reasoning process on task-oriented dialogue summarization.

\section{Results}

\subsection{Quantitative Analysis} \label{sec:quantitative_analysis}
\paragraph{Comparison with SotA methods.}
\leavevmode\newline
We begin our analysis using standard metrics capturing both N-gram overlap (R1, R-2, R-L and BLEU) and semantic similarity (BERT-Score). While we acknowledge that these metrics alone cannot fully assess summary quality, they provide a useful foundation for further in-depth analysis.

\noindent{\large \textcircled{\small 1}} SAMSum: Table~\ref{tab:all_results} (top) shows that \app outperforms both SotA supervised and unsupervised methods, achieving relative improvements of +19\% BLEU and +3\% BERT-Score w.r.t.\ \citet{tian2024dialogue}. Among baselines, DeepSeek-R1-Distill-14B performs the best, slightly ahead of others. The top summarizer (Llama-3.1-8B-Instruct) achieves +34\% BERT-Score improvement after \app's third step. Table~\ref{tab:generation_samsum} shows the different summarizers' performance on this dataset for the First Step (Generation) and illustrates the performance boost from \app's Step (Evaluation) through collaborative selection. 

\noindent{\large \textcircled{\small 2}} DialogSum: While all baselines achieve comparable results, DeepSeek-R1-Distill-14B emerges as the best among them with a BERT-Score of 55.47\%. \app outperforms both the baselines and the previous SotA MoE \citep{tian2024dialogue} by achieving a BERT-Score of 68.72\%$\pm 0.44$.

\noindent{\large \textcircled{\small 3}} MTS-Dialog: 
Here again the best summarizer is Llama-3.1-8B-Instruct and its performance improves by 31\%, 68\% and 9\% in terms of Rouge-L, BLEU, and BERT-Score respectively. Notably, it surpasses BART-GS-DA \citep{abacha2023empirical} by 50\% BERT-Score while outperforming other unsupervised baselines, including GPT-3, Llama, and DeepSeek.

\begin{table}[h]
    \scalebox{0.74}{
    \begin{tabular}{lccccc}
        
        \textbf{Summaries} & \textbf{R-1} & \textbf{R-2} & \textbf{R-L} & 
        \textbf{BLEU} & \textbf{BERT-Score}  \\ \toprule
        \multicolumn{6}{c}{Generation (First Step)} \\ 
         {Lama 3.1} & \textbf{44.95} & 19.45 & \textbf{35.40} & \textbf{12.34} & \textbf{67.69} \\ 
        {Gemma 2}   & 43.72 & 18.33 & 34.40 & 11.07 & 67.05 \\ 
        {Qwen 2}    & 44.15 & \textbf{19.55} & 35.16 & 11.64 & 67.65 \\ 
        \midrule
        \multicolumn{6}{c}{Evaluation (Second Step)} \\ 
         {Best Selected} & \textbf{50.43} & \textbf{25.06} & \textbf{41.11} & \textbf{15.82} & \textbf{71.59} \\ \bottomrule
                        
    \end{tabular}
    }
    \caption{Evaluation of the generated and the selected summaries on the SAMSum dataset.}
    \label{tab:generation_samsum}
\end{table}

\begin{table*}[h]
    \centering
    \scalebox{0.9}{
    \begin{tabular}{lcccccc}
        \toprule
        \textbf{Setting} &  Vanilla & 0-Shot \app & 10\% Sup. & \app & \app + 10\% Sup. & Full Sup. \\ \midrule
        \textbf{BERT-Score} & 53.70 & 65.99 & 67.10 & 68.72 & 73.02 & 73.06\\ 
        \textbf{Rouge-L}        & 16.46 & 29.16 & 29.38 & 32.68 & 37.80 & 37.73 \\ \bottomrule        
    \end{tabular}
    }
    \caption{Impact of supervision variants on \app. \textbf{``Sup.''} 
    denotes the ratio of supervised training data used.}
    \label{tab:quartz_vs_supervision}
\end{table*}

\paragraph{The impact of incorporating supervision.}
While \app can achieve superior performance to supervised methods (see Table~\ref{tab:all_results}), it can also leverage limited supervision (10\% of the training set) through LoRA fine-tuning (\textit{\app + 10\% Sup.}) to reach near full-supervision results (Table~\ref{tab:quartz_vs_supervision}).
In contrast, training the same baseline (Llama-3.1-8B-Instruct) on the same 10\% data fails to exceed a BERT-Score of 68\%. Notably, \textit{0-Shot \app} (i.e., without fine-tuning), surpasses the \textit{Vanilla} baseline by $+23\%$ BERT-Score and $+77\%$ Rouge-L, demonstrating robustness in low-resource or fine-tuning-limited scenarios.

\begin{figure}[ht!]
\centering
    \includegraphics[width=0.48\textwidth]{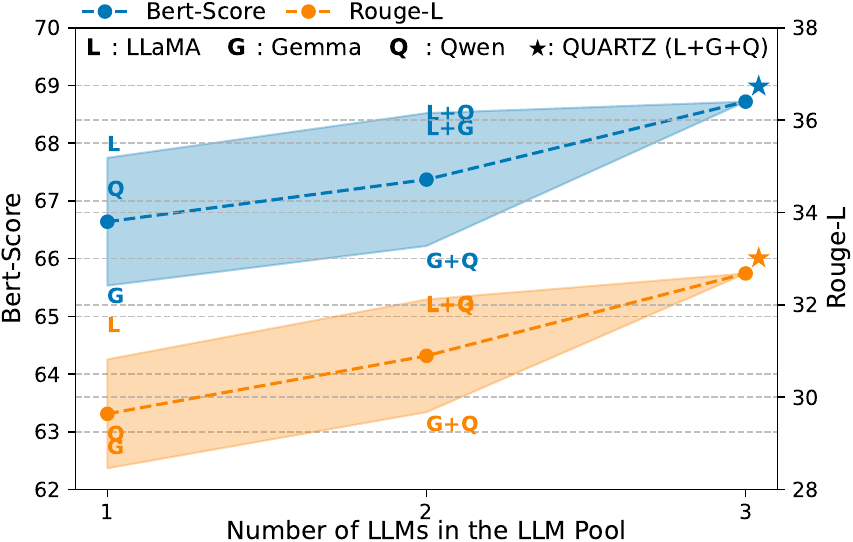}
    \caption{Impact of LLM pool size on summarization performance. \textbf{\textit{Dotted lines}} represent the mean performance. \textbf{\textit{Shaded regions}} indicate the standard deviation across the combinations (see Appendix~\ref{sec:appendix_llm_pool_config}).} 
    \label{fig:num_llms}
\end{figure}

\paragraph{The benefits of a diverse LLM pool over a single model.}
Recent work \citep{subramaniam2025multiagent} highlights that using a pool of LLMs promotes diversity and fosters richer model interaction. This is clearly demonstrated in Table~\ref{tab:generation_samsum} where the selected summaries using the LLM pool outperform those generated by each individual model in the pool across all metrics. While Llama-3.1-8B-Instruct generated $58\%$ of the best summaries across all datasets, the contributions of other models are still significant (see Appendix~\ref{sec:appendix_win_rate}, Figure~\ref{fig:win_rate}). Figure~\ref{fig:num_llms} shows that increasing the LLM pool size $(|L|)$ enhances summary diversity and performance, though with higher computational cost (see Appendix~\ref{sec:appendix_llm_pool_config}).

\begin{table*}[h]
    \centering
    \small
    \scalebox{0.9}{
    \begin{tabular}{p{0.49\linewidth}|p{0.37\linewidth}|p{0.15\linewidth}}
        \toprule
         \multicolumn{1}{c}{\textbf{Zero-Shot \app (First Step + Second Step Only)}} &  \multicolumn{1}{c}{\textbf{\app (All Three Steps)}} & \multicolumn{1}{c}{\textbf{Note}} \\ \midrule

        {Possible psychiatric disorders or conditions that may be causing the patient's behavior, such as a psychotic disorder.} & {Psychotic disorder, \textbf{possibly schizophrenia or bipolar disorder}.} & {\hlgreen{Conciseness, Specificity}} \\ \midrule

        {Diagnostic Hypotheses: Self-inflicted stab wound.} & {Diagnostic Hypotheses: Self-inflicted stab wound, \textbf{potential suicidal intent}} & {\hlgreen{Flags suicidal intent}} \\ \midrule

        {The patient's fourteen-year-old daughter is experiencing abnormal behavior, including difficulty speaking and walking, after a fight with her mother. \textbf{The daughter smells of alcohol.}} & {Altered mental status and \textbf{ataxia} in a 14-year-old female following a fight with her mother.} & {\hlpink{Lacks alcohol detail}, \hlgreen{Includes medical terminology}} \\ \midrule

        {The patient is the neighbor of the individual requiring medical attention. \textbf{The call was made from the neighbor's home}.} & {The caller is the patient's neighbor, \textbf{calling from the patient's home}.} & {\hlpink{Incorrect Call Context}}  \\

        \bottomrule    
    \end{tabular}
    }
    \caption{Impact of \app's Third Step (Fine-Tuning) on model alignment. \textbf{Left}: Zero-Shot \app generates multiple summaries and selects the best ones. \textbf{Right}: Additional fine-tuning is applied to the selected summaries} 
    \label{tab:before_after_simsamu}
\end{table*}

\subsection{Qualitative Analysis}
While \app promotes factually consistent and task-relevant summaries via QA-based evaluation, standard metrics (Section~\ref{sec:quantitative_analysis}) fall short of capturing this. We therefore conduct a qualitative analysis to examine how \app improves factual consistency and task relevance.

\paragraph{Enhancing task relevance and factual soundness.}

The selected summaries consistently incorporate more task-relevant information, driven by task-specific QA generation and factual consistency throughout the ranking process of the candidate answers. These selected summaries often introduces new insights absent in other. Examples are provided in Appendix~\ref{sec:appendix_simsamu_summaries}, Figures~\ref{fig:generation_exemple1_dialogsum},~\ref{fig:generation_exemple2_dialogsum} and~\ref{fig:generation_exemple_samsum}.

\paragraph{Impact of fine-tuning on summarization quality.}
\leavevmode\newline 
On the SimSAMU dataset, the medical expert involved in this study noted that fine-tuned summaries were clearer and directly to the point compared to those generated before fine-tuning. This is particularly valued by emergency regulators in time-sensitive situations. Table~\ref{tab:before_after_simsamu} shows key enhancements and potential degradations observed 

(more examples in Appendix~\ref{sec:appendix_simsamu_summaries} and Table~\ref{tab:before_after_simsamu_long}).

\section{LLM as a Judge}
\label{sec:llm_judge}
Motivated by recent findings that LLMs can serve as reliable proxies for human annotators \citep{song2024finesure}, we adopt the G-Eval framework \citep{liu2023g} using two top-ranked  open-weights judge models from the Judge Arena \citep{zheng2023judging}: \textit{Selene-1-Mini-Llama-3.1-8B} \cite{alexandru2025atlaseleneminigeneral} and \textit{Llama-3.3-70B-Instruct} \cite{dubey2024llama}, to ensure reproducibility, contrary to the original GPT4-based setup. Following \citet{liu2023g} we consider the coherence (COH), consistency (CON), fluency (FLU) and relevance (REL) dimensions (as detailed in Appendix \ref{sec:appendix_llm_judge}). As shown in Table~\ref{tab:llm_judge_metrics}, both judges consistently assign lower scores to reference summaries, suggesting that LLM-generated summaries can already provide strong initial candidates. Although the judge models assign different absolute scores, their relative rankings are globally preserved. The 0-Shot configuration (\app: Step 1 + Step 2) identifies summaries that perform well on individual metrics (e.g., CON, FLU), but finding candidates that excel across all dimensions remains challenging due to trade-offs in the LLMs-generated summaries set. Fine-tuning addresses this by steering generation toward better overall trade-offs.

\begin{table}[h]
    \scalebox{0.53}{
    \begin{tabular}{lccccc|ccccc}
        \toprule
        \multirow{3}{*}{\bf \diagbox[width=5.5em]{Config}{Judge}} & \multicolumn{5}{c|}{\bf Selene-1-Mini-Llama-3.1-8B} & \multicolumn{5}{c}{\bf Llama-3.3-70B-Instruct}\\ \cmidrule(lr){2-11}

        & \textbf{COH} & \textbf{CON} & \textbf{FLU} & \textbf{REL} & \textbf{Avg} & \textbf{COH} & \textbf{CON} & \textbf{FLU} & \textbf{REL} & \textbf{Avg} \\ \midrule

        \textbf{Llama 3.1}     & 3.79 & 4.12 & 3.54 & 3.86 & 3.82   & 4.26 & 4.34 & 3.92 & 4.29 & 4.20 \\
        \textbf{Gemma}         & 3.77 & 4.13 & 3.62 & \textbf{4.04} & \textbf{3.89}   & \textbf{4.41} & 4.39 & \textbf{4.12} & \textbf{4.52} & \textbf{4.36} \\
        \textbf{Qwen}          & \textbf{3.88} & 4.12 & 3.63 & 3.96 & \textbf{3.89}   & 4.29 & 4.37 & 3.98 & 4.41 & 4.26 \\
        \textbf{0-Shot}        & 3.86 & \textbf{4.14} & \textbf{3.66} & 3.86 & 3.88   & 4.28 & \textbf{4.40} & 4.00 & 4.36 & 4.26 \\
        \textbf{Reference}     & 3.47 & 3.91 & 3.31 & 3.52 & 3.55   & 3.89 & 3.92 & 3.60 & 3.86 & 3.81 \\ \midrule
        \textbf{\app}          & \textbf{3.89} & \textbf{4.18} & \textbf{3.66} & 3.87 & \textbf{3.90}   & \textbf{4.47} & \textbf{4.41} & 3.96 & \textbf{4.53} & \textbf{4.34} \\
        
    \end{tabular}
    }
    \caption{LLM-as-judge evaluation scores for summaries using two models. COH, CON, FLU, REL, and Avg denote coherence, consistency, fluency, relevance, and average score, respectively.}
    \label{tab:llm_judge_metrics}
\end{table}

\section{Human Evaluation}
\label{sec:human_eval}

Given the high cost of human evaluation, we adopted a two-phase protocol with four expert annotators: all computer scientists, including one with a medical degree to ensure informed assessment, particularly for clinical dialogues. 

\textbf{Phase One.} Using the Potato annotation tool \cite{pei2022potato}, annotators reviewed each dialogue along with two anonymized summaries, one generated by \app and the other being the reference. Without knowing the source of either summary, they were asked to choose the one they preferred based on overall quality. Notably, in 48\% of cases, the annotators favored \app summaries over the reference with a Fleiss' kappa \cite{fleiss1973equivalence} agreement of 0.14 (slight agreement). Although no "equal quality" option was provided, some annotators reported difficulty in choosing a winner, stating that both summaries were often equally informative.
\textbf{Phase Two.} Annotators evaluated the final summaries produced by \app according to the four qualitative dimensions as in Section \ref{sec:llm_judge}. The average scores are presented in Table~\ref{tab:human_eval}, with annotators reaching 39.5\% inter-annotator agreement across all evaluation criteria. 

\begin{table}[h]
    \centering
    \scalebox{0.7}{
    \begin{tabular}{lccccc}
        \toprule
        & \textbf{COH} & \textbf{CON} & \textbf{FLU} & \textbf{REL} & \textbf{Avg} \\ \midrule
        \textbf{\app} (1–5 Likert) & 4.17 & 4.01 & 4.27 & 4.06 & 4.12 \\ \midrule
        \textbf{Cohen's kappa} ([-1; 1]) & 0.12 & 0.17 & 0.14 & 0.09 & 0.13 \\
        \textbf{Exact Agreement} (\%) & 42 & 42 & 40 & 34 & 39.5 \\
        
    \end{tabular}
    }
    \caption{Human evaluation for \app final summaries and Cohen's kappa and exact agreement rates.}
    \label{tab:human_eval}
\end{table}

\section{Conclusion and Future Work}
We introduced \app, a framework that harnesses a pool of LLMs for unsupervised dialogue summarization. \app begins by generating diverse summaries and task-related QA pairs, then employs a two-stage evaluation process to identify the most informative summaries based on answer quality.
Finally, \app fine-tunes the top-performing LLM summarizer on the selected generated summaries. We evaluate it using statistical and embedding-based scorers, LLM as a judge and human evaluation. Experiments across multiple domains demonstrate the effectiveness of \app, consistently surpassing state-of-the-art supervised methods. We believe \app has broad practical value, particularly in domains like healthcare and meeting summarization, where it can help professionals efficiently structure key information and support tasks such as clinical documentation and entity extraction.
For future work, we plan to unveil the impact of \textit{iterative} \app (multi-iteration LLM selection and specialization) on summary quality, and explore other fine-tuning techniques to tackle the pool of LLMs in a different fashion.

\section{Limitations}
Although we focused on minimizing data annotation costs while maintaining high-quality summaries, some limitations remain. For the sake of simplicity, we did not act on the generated gold question-answer pairs. While we can explicitly incorporate questions relevant to the task, we think that it would be beneficial to establish some control over the questions based on their contribution to summary evaluation, as not all question-answer pairs carry the same level of importance. While \app is designed for unsupervised summarization, the lack of human oversight during training and its reliance on LLM performance may lead to discrepancies from human-preferred summaries or overlook subtle errors. Although human evaluations provided high scores across all dimensions (average ratings over 4 on a 1-5 Likert scale), inter-annotator agreement was low, indicating that even expert judgments might differ substantially. The employment of supplementary evaluation techniques is encouraged by this, which highlights the difficulties in assessing subjective components of summary quality.

\section{Ethical Considerations}

While this work demonstrates remarkable advancements toward reference-free dialogue summarization, it is important to be aware of the limitations and risks of relying on such approaches in particularly sensitive and high-stakes areas, especially in healthcare. Generative models present serious challenges, such as implicit assumptions embedded in their training data and difficulties in generalizing across wide and complex domains. These limitations will need drastic scrutiny when applying such models in high-stakes environments.  For our  triage-oriented summarization task, each decision made by the model was checked by a medical expert to ensure its accuracy and relevance. However, the integration of automated systems in healthcare raises ethical and legal concerns regarding accountability and privacy. These considerations further validate our decision to operate on open-source models. Unlike closed-source models, which may involve transferring sensitive private data to third-party companies and expose it to potential misuse, open-source approaches allow for greater transparency and control over the underlying data and processes. Thus, while LLMs have shown promising capabilities, their application in critical fields must be approached with caution to ensure that ethical standards are maintained, and that these systems do not inadvertently compromise patient safety or privacy.

\section{Acknowledgments}

This work was supported by the French National Research Agency (ANR) under the project LLM4All (ANR-23-IAS1-0008).

\bibliography{custom}
\appendix

\section{LLM Pool Setup and Selection Dynamics}
\subsection{Details on LLM Pool Configurations} \label{sec:appendix_llm_pool_config}
We presented in Figure~\ref{fig:num_llms} how the number $|L|$ of LLMs in the LLM pool impacts summarization quality. The LLMs are chosen among these 3 models: Llama-3.1-8B-Instruct, Gemma-2-9b-it, and Qwen2-7B-Instruct. For each pool size $|L|$ we evaluate all $\mathcal{C}^{|L|}_{3}$ possible model combinations. In Figure~\ref{fig:num_llms}, dotted lines represent the mean values, while shaded regions indicate the standard deviation. For $|L| = 3$, there is only one possible combination, which corresponds to the actual \app configuration reported in the last row of Table~\ref{tab:all_results} for the DialogSum dataset. 

\subsection{Prioritizing the Best LLM Summarizer for Fine-Tuning}
\label{sec:appendix_win_rate}
\begin{figure}[ht!]
\centering
    \includegraphics[width=0.47\textwidth]{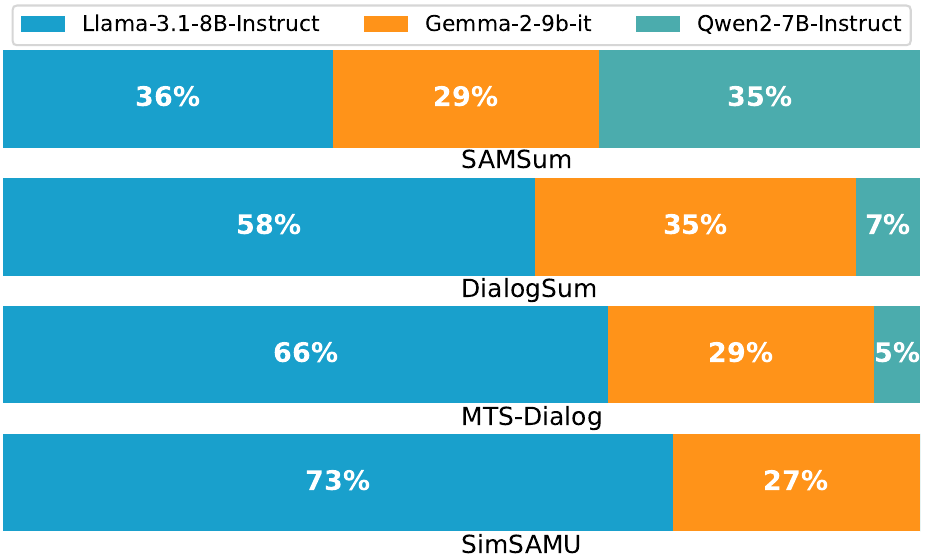}
    \caption{Win Rate of LLMs Across Datasets} 
    \label{fig:win_rate}
\end{figure}

{\app}’s final step (see Section~\ref{sec:finetune}) aims to fine-tune the best LLM summarizer, i.e., the model that generated the majority of top-selected summaries. Figure~\ref{fig:win_rate} illustrates the proportion of best-selected summaries across datasets, with Llama-3.1-8B-Instruct contributing 58\%, Gemma-2-9b-it 30\%, and Qwen2-7B-Instruct 12\%. Empirical findings indicate that fine-tuning the top-performing model yields the most effective summarization results, outperforming alternative selections for the final step.

\section{Implementation Details}
\label{sec:appendix_implementation_details}

We fine-tune the LLM summarizer using LoRA \citep{hulora} for 3 epochs with rank $r_\text{LoRA}=8$ and scaling factor $\alpha_\text{LoRA}=16$. Optimization is performed using the AdamW optimizer with a learning rate of $5\times10^{-5}$, $\beta_1=0.9$, $\beta_2=0.999$, and $\epsilon=1\times10^{-8}$. A linear learning rate scheduler is applied. All experiments were conducted on a single A100-40GB GPU. For full \app pipeline (comprising the 3 steps), the total computation time amounts to 11 GPU hours. Once the best summarizer is obtained, it can be used for online inference and further optimized via inference optimization toolkits.

\section{Annotator Guidelines}
\label{sec:annotator_guidelines}

To ensure reliable human evaluation, we conducted a two-phase annotation process with four expert annotators. In \textbf{Phase 1}, annotators were presented with a dialogue and two anonymized summaries (one from \app and one from the reference) and asked to select the preferred one. In \textbf{Phase 2}, annotators independently rated each \app’s summary along four quality dimensions—\textit{Coherence}, \textit{Consistency}, \textit{Fluency}, and \textit{Relevance} on a 5-point Likert scale. The criteria for scoring were as follows:

\begin{itemize}
    \item \textbf{Coherence (1–5):} Assesses the overall logical flow and structure of the summary.

    \item \textbf{Consistency (1–5):} Evaluates factual alignment with the source dialogue.

    \item \textbf{Fluency (1–5):} Rates grammaticality, readability, and linguistic quality.

    \item \textbf{Relevance (1–5):} Measures how well the summary captures important content without redundancy.
\end{itemize}

\section{Prompt Templates}
\label{sec:appendix_prompts}
\subsection{Summary Generation}

\noindent \textbf{SAMSum:}

\noindent \texttt{\textbf{``instruction'':} You will be provided with a conversational exchange that simulates a natural messaging or chat-like interaction.
Your task is to produce a short, concise and clear summary that captures the most important points and key information conveyed throughout the exchange.}

\noindent \texttt{\textbf{``input'':} [Conversation]}

\vspace{12pt} \noindent \textbf{MTS-Dialog:}

\noindent \texttt{\textbf{``instruction'':} You are a medical scribe tasked with writing concise yet informative medical notes based on doctor-patient interactions. Your goal is to create clear and professional note text about from the patient doctor dialogue focusing on [Header]}

\noindent \texttt{\textbf{``input'':} [Conversation]}

\vspace{12pt} \noindent \textbf{SimSAMU:}

\noindent Prompt design for this triage-related dataset was assisted by a medical expert to ensure that the generated notes are contextually relevant and support accurate triage decisions during critical incidents.

\noindent \texttt{\textbf{``instruction'':} You are an experienced emergency physician handling a telephone consultation. Your task is to summarize the following medical dialogue into a precise and structured clinical report.
When summarizing, ensure the following:
- Use concise, clear, and professional language.
\\- Translate informal or everyday terms into appropriate medical terminology where possible.
\\- Maintain the structure provided below, leaving any sections blank if the required information is unavailable.
Format of the clinical report:
\\Complete the following sections in order.
\\1-Chief Complaint: The main medical issue prompting the call (e.g., chest pain).
\\2-Call Context: The relationship between the caller and the patient (e.g., patient themselves, spouse, bystander) and the location of the call (e.g., home, street).
\\3-Patient Context: Demographic information (age, sex), social situation (e.g., lives alone, resides in a retirement home), and degree of autonomy.
\\4-Usual Treatment: Current medications or treatments for known comorbidities.
\\5-Past Medical History: Relevant medical, allergic, or surgical history (e.g., diabetes, prior surgeries).
\\6-Patient Symptoms: Symptoms as reported, categorized into: 
\\- General symptoms (e.g., fever, fatigue).
\\- Organ-specific symptoms (e.g., respiratory: shortness of breath).
\\7-History of Present Illness: A detailed, chronological account of the events leading to the call, describing their sequence and interrelations (free text).
\\8-Diagnostic Hypotheses: Possible diagnoses based on the information provided (e.g., myocardial infarction).
\\9-Treatment Plan: Recommendations including medications, therapies, lifestyle advice, referrals, or additional diagnostic tests.
\\10-Triage Decision: The proposed course of action (e.g., remain at home, dispatch a doctor, self-transport to the emergency department, send an ambulance, or dispatch a medicalized ambulance)}
\noindent \texttt{\textbf{``input'':} [Conversation]}

\subsection{QAs Generation}
\noindent \textbf{SimSAMU:}

\noindent \texttt{\textbf{``instruction'':}Given this emergency call between a doctor and a patient, generate 10 to 15 question and answer pairs that are relevant to the medical triage and that should be present in the clinical record.
\\Do not repeat the same question or answer.
\\Do not ask questions that you can't answer based on the information provided in the dialogue.
\\Format the questions and answers as follows: Q1: <question1> A1: <answer1> }

\noindent \texttt{\textbf{``input'':}[Conversation]}

\subsection{Response Evaluation}

\noindent \texttt{\textbf{``instruction'':} You will be provided with a ground truth answer and a list of generated answers. Your task is to rank the generated answers based on their correctness and closeness to the ground truth answer. The ground truth answer appears first, followed by the list of generated answers. The ranking should be a three-element list of integers between 1 and 3, where 1 represents the most accurate answer and 3 represents the least accurate. If an answer is NOT\_INCLUDED, place it at the end of the ranking.}
\noindent \texttt{\textbf{``input'':} Question: [Question]
\\Ground Truth Answer: [Ground Truth Answer]
\\Possible answers:
\\1) [Answer\_1]
\\2) [Answer\_2]
\\3) [Answer\_3]}

\subsection{LLM as a Judge}
\label{sec:appendix_llm_judge}
\noindent \textbf{Coherence metric:}

\noindent \texttt{\textbf{``instruction'':} You are a helpful assistant tasked with evaluating how coherent a summary is for a given dialogue. Your goal is to rate the summary based only on how well its sentences form a clear, logical, and well-structured presentation of the dialogue content.
Assign a score from 1 to 5 based solely on *coherence*:
\\5: Excellent - the summary is well-organized, easy to follow, and logically structured.
\\4: Good - mostly coherent with only minor issues in flow or structure.
\\3: Fair - somewhat coherent but with noticeable issues in clarity or organization.
\\2: Poor - disorganized, unclear, or hard to follow.
\\1: Very poor - sentences feel disconnected or incoherent, severely impacting understanding.
\\Only reply with the number **1**, **2**, **3**, **4**, or **5**. Do not include any explanation or extra text.
\\Your reply should strictly follow this format:  
**Score:** <1, 2, 3, 4, or 5>}
    
\noindent \texttt{\textbf{``input'':} Dialogue: [Dialogue]
\\Summary: [Summary]}

\noindent \textbf{Consistency metric:}

\noindent \texttt{\textbf{``instruction'':} You are a helpful assistant tasked with evaluating how factually consistent a summary is with a given dialogue. Your goal is to rate the summary based only on whether it accurately reflects the facts stated in the original dialogue without introducing unsupported or hallucinated information.
Assign a score from 1 to 5 based solely on *consistency*:
\\5: Excellent — all statements in the summary are fully supported by the dialogue.
\\4: Good — minor inaccuracies or slight overgeneralizations, but mostly faithful to the dialogue.
\\3: Fair — some factual inconsistencies or minor hallucinations are present.
\\2: Poor — several statements in the summary are not supported or contradict the dialogue.
\\1: Very poor — the summary contains major hallucinations or is largely inconsistent with the dialogue.
\\Only reply with the number **1**, **2**, **3**, **4**, or **5**. Do not include any explanation or extra text.  
\\Your reply should strictly follow this format:  
**Score:** <1, 2, 3, 4, or 5>}
    
\noindent \texttt{\textbf{``input'':} Dialogue: [Dialogue]
\\Summary: [Summary]}

\noindent \textbf{Fluency metric:}

\noindent \texttt{\textbf{``instruction'':} You are a helpful assistant tasked with evaluating how fluent a summary is for a given dialogue. Your goal is to rate the summary based only on grammar, spelling, punctuation, word choice, and sentence structure.
Assign a score from 1 to 5 based solely on *fluency*:
\\5: Excellent — the summary is free of errors and reads very smoothly.
\\4: Good — the summary has minor errors but is easy to read.
\\3: Fair — the summary has some noticeable errors that slightly affect clarity or flow.
\\2: Poor — the summary has many errors that affect understanding or naturalness.
\\1: Very Poor — the summary contains frequent errors making it difficult to understand.
\\Only reply with the number **1**, **2**, **3**, **4**, or **5**. Do not include any explanation or extra text.  
\\Your reply should strictly follow this format:  
**Score:** <1, 2, 3, 4, or 5>}
    
\noindent \texttt{\textbf{``input'':} Dialogue: [Dialogue]
\\Summary: [Summary]}

\noindent \textbf{Relevance metric:}

\noindent \texttt{\textbf{``instruction'':} You are a helpful assistant tasked with evaluating how relevant a summary is for a given dialogue. Your goal is to rate the summary based only on how well it includes important information from the dialogue while avoiding irrelevant or redundant content.
Assign a score from 1 to 5 based solely on *relevance*:
\\5: Excellent — the summary covers all key points and contains no irrelevant or redundant information.
\\4: Good — the summary covers most key points with minimal irrelevant or redundant content.
\\3: Fair — the summary covers some key points but includes noticeable irrelevant or redundant information.
\\2: Poor — the summary misses many key points and contains significant irrelevant or redundant content.
\\1: Very Poor — the summary largely fails to capture key points and is mostly irrelevant or redundant.
\\Only reply with the number **1**, **2**, **3**, **4**, or **5**. Do not include any explanation or extra text.  
\\Your reply should strictly follow this format:  
**Score:** <1, 2, 3, 4, or 5>}
    
\noindent \texttt{\textbf{``input'':} Dialogue: [Dialogue]
\\Summary: [Summary]}

\section{Example \app Summaries}

\label{sec:appendix_simsamu_summaries}
\noindent \textbf{SimSAMU:}
\\We use \hlpink{RED} to highlight information present in the pre-fine-tuning summary but omitted after fine-tuning, and \hlgreen{GREEN} to indicate information added in the post-fine-tuning summary, in order to visualize differences between the two model outputs.

\noindent \textbf{Example 01:}
 \\ \noindent \texttt{1-Chief Complaint: Self-inflicted stab wound.
\\2-Call Context: The patient's cousin called for the brother, who is the patient. The call was made from the patient's home.
\\3-Patient Context: Age and sex are unavailable. The patient lives at home with his brother. The degree of autonomy is unclear.  \hlpink{It is mentioned that the patient's brother is in a residence where the patient resides.}
\\4-Usual Treatment: Unavailable.
\\5-Past Medical History: Unavailable.
\\6-Patient Symptoms: General symptoms: \hlgreen{agitation, refusal to seek help}. Organ-specific symptoms: unavailable.
\\7-History of Present Illness: The patient's brother was found with blood on his pants after the patient saw him stab himself again. The brother refused to let the patient look at the wound and did not want him to call for help. \hlgreen{The patient's cousin called for help instead}. The patient was reluctant to involve the authorities, fearing his brother's reaction.
\\8-Diagnostic Hypotheses: Self-inflicted stab wound, \hlgreen{potential suicidal intent}.
\\9-Treatment Plan: \hlgreen{Dispatch of firefighters to the scene to handle the situation.} The patient is advised to stay with his brother and remind the authorities of any new developments.
\\10-Triage Decision: \hlpink{The emergency services decided to} Dispatch of firefighters\footnote{Firefighters are commonly responsible for responding to medical emergencies in France.} to the scene  \colorbox{pink}{to handle the situation}.}

\noindent \textbf{Example 02:}
\\ \noindent \texttt{1-Chief Complaint: Malaise, \hlgreen{possibly a vagal malaise}, \hlpink{episode of feeling like leaving without losing consciousness}, \hlgreen{characterized by a feeling of impending syncope without actual loss of consciousness}.
\\2-Call Context: The patient is the daughter of the 85-year-old man experiencing the malaise, calling from her home.
\\3-Patient Context: The patient is an 85-year-old man with \hlgreen{hypertension}, living with his daughter, who is an emergency worker.
\\4-Usual Treatment: He takes treatments for hypertension.
\\5-Past Medical History: Hypertension.
\\6-Patient Symptoms: 
General symptoms: Malaise, \hlgreen{feeling of impending syncope}.
Organ-specific symptoms: None mentioned.
\\7-History of Present Illness: The patient experienced a malaise while in the bathroom, characterized by a feeling of impending syncope. He called his daughter, \hlpink{who assisted him and he recovered.}, \hlgreen{who arrived and turned him into a supine position with his legs elevated, which improved his condition}.
\\8-Diagnostic Hypotheses: Vagal malaise.
\\9-Treatment Plan: The patient should be taken to the emergency room for further evaluation. He should bring his prescription for his usual treatments.
\\10-Triage Decision: The patient should be transported to the emergency room \hlgreen{by his daughter}, who should ensure his comfort during the transport.}

\bigskip
\bigskip

\begin{figure*}[ht!]
\centering
    \includegraphics[width=\textwidth]{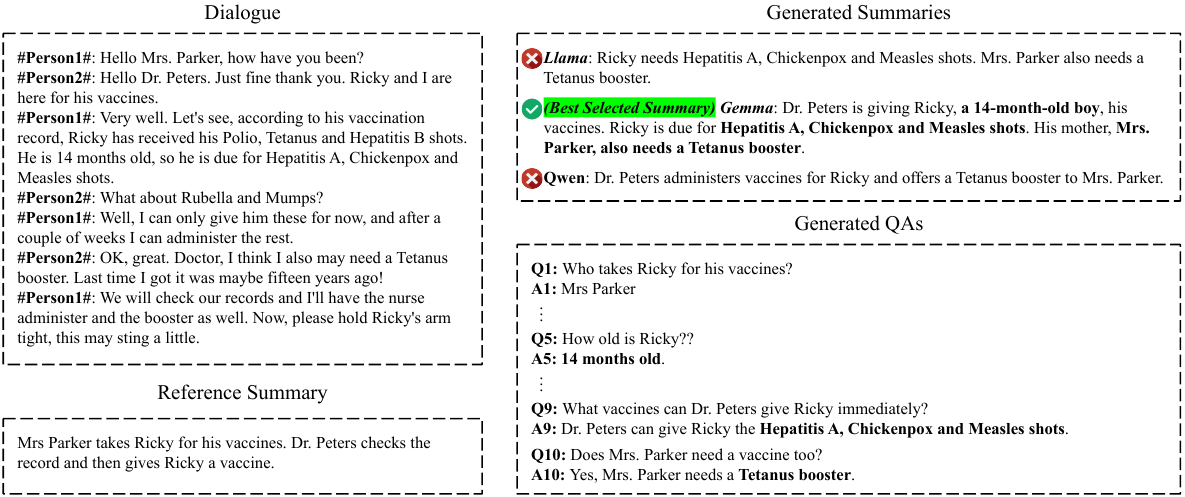}
    \caption{An example dialogue from the DialogSum dataset. \textbf{Left:} The original dialogue and the reference summary. \textbf{Right:} The generated summaries from each LLM in the pool along with the task-related QAs. Additional information introduced by the best-selected summary is highlighted in \textbf{bold}.} 
    \label{fig:generation_exemple1_dialogsum}
\end{figure*}

\noindent \textbf{DialogSum:}
\\In Figure~\ref{fig:generation_exemple1_dialogsum}, we present a dialogue example alongside the summaries and question-answer pairs generated by LLMs. Notably, the best-selected summary (LLM Summarizer: Gemma-2-9b-it) incorporates additional facts beyond both the generated and reference summaries. While this enhances task-oriented summarization by enriching the information provided, it is undervalued by surface-level metrics.

\begin{figure*}[ht!]
\centering
    \includegraphics[width=\textwidth]{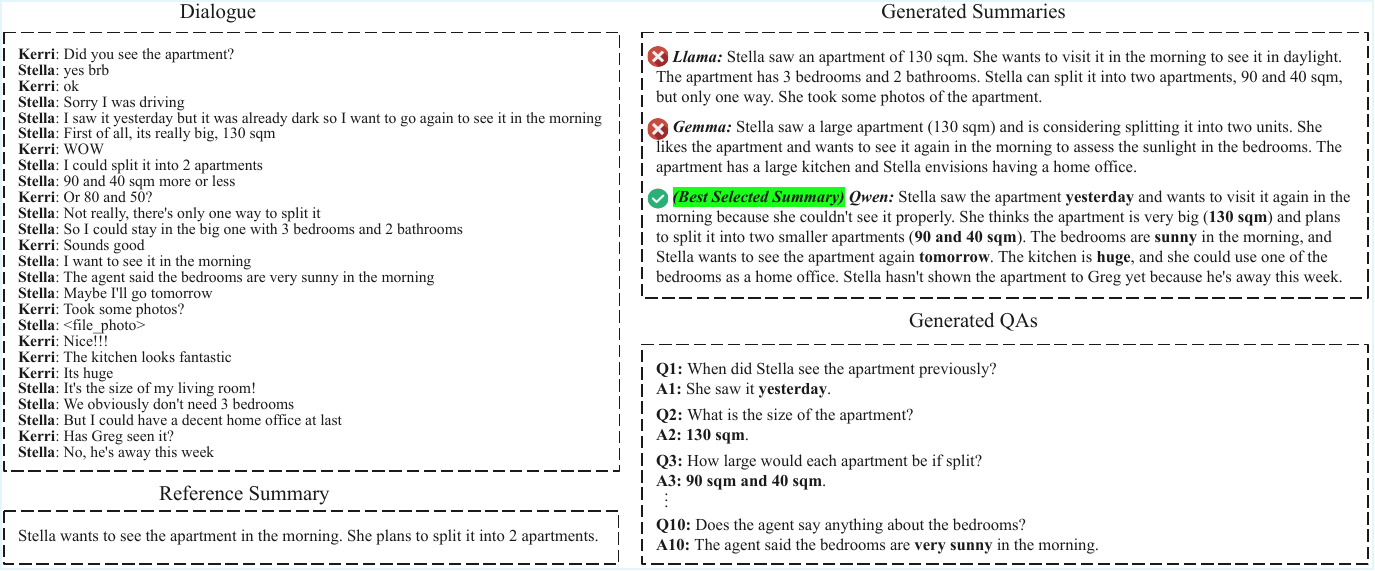}
    \caption{Another example dialogue from the DialogSum dataset. \textbf{Left:} The original dialogue and the reference summary. \textbf{Right:} The generated summaries from each LLM in the pool along with the task-related QAs. Additional information introduced by the best-selected summary is highlighted in \textbf{bold}.} 
    \label{fig:generation_exemple2_dialogsum}
\end{figure*}

\begin{figure*}[ht!]
\centering
    \includegraphics[width=\textwidth]{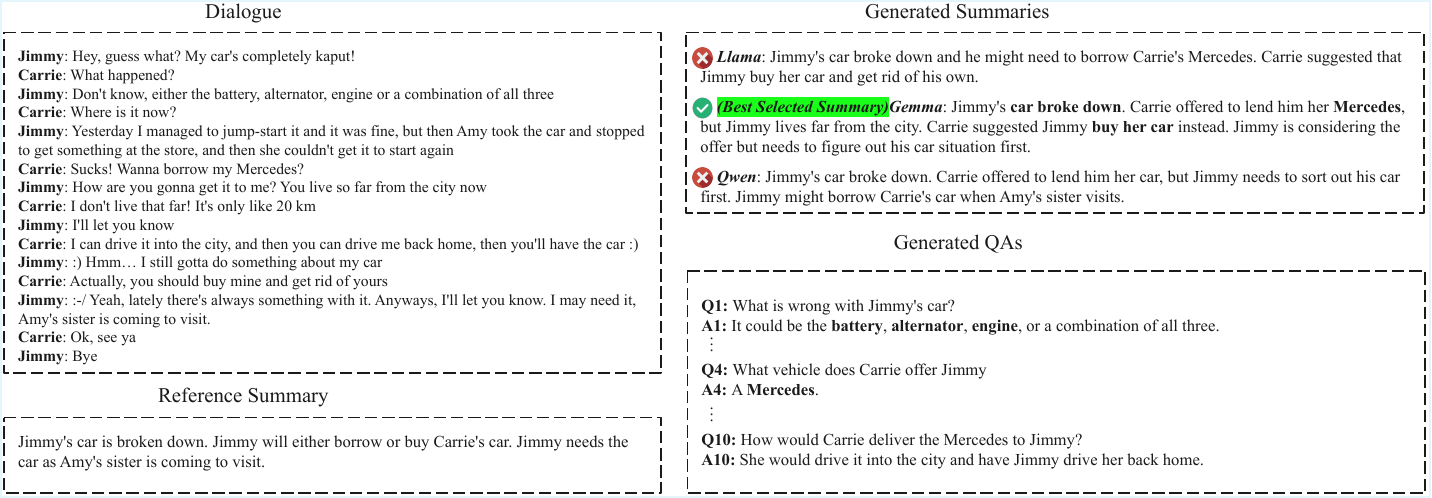}
    \caption{Another example dialogue from the SAMSum dataset. \textbf{Left:} The original dialogue and the reference summary. \textbf{Right:} The generated summaries from each LLM in the pool along with the task-related QAs. Additional information introduced by the best-selected summary is highlighted in \textbf{bold}.} 
    \label{fig:generation_exemple_samsum}
\end{figure*}

\begin{table*}[h]
    \centering
    \small
    \scalebox{0.9}{
    \begin{tabular}{p{0.48\linewidth}|p{0.37\linewidth}|p{0.15\linewidth}}
        \toprule
         \multicolumn{1}{c}{\textbf{Zero-Shot \app (First Step + Second Step Only)}} &  \multicolumn{1}{c}{\textbf{\app (All Three Steps)}} & \multicolumn{1}{c}{\textbf{Note}} \\ \midrule

        {Possible psychiatric disorders or conditions that may be causing the patient's behavior, such as a psychotic disorder.} & {Psychotic disorder, \textbf{possibly schizophrenia or bipolar disorder}.} & {\hlgreen{Conciseness, Specificity}} \\ \midrule

        {Diagnostic Hypotheses: Self-inflicted stab wound.} & {Diagnostic Hypotheses: Self-inflicted stab wound, \textbf{potential suicidal intent}} & {\hlgreen{Flags suicidal intent}} \\ \midrule

        {The patient's husband called for his wife, who is experiencing a medical issue.} & {The patient's husband called for his wife, who is being \textbf{followed by a psychiatrist for depression}.} & {\hlgreen{Adds context}} \\ \midrule
                 
        {The patient is experiencing malaise and has had a fall, with symptoms including a spinning head and loss of balance.} & {Malaise, \textbf{syncope (fainting)}, and a fall.} & {\hlgreen{Includes medical terminology}} \\ \midrule

        {The patient also has a sore throat since the morning and a pain in the left arm.} & {The pain is located in the \textbf{upper abdomen and radiates to the left arm}.} & {\hlgreen{Precise anatomical detail}} \\ \midrule
        
        {Possible myocardial infarction, considering the chest pain and arm pain.} & {Possible diagnoses include peptic ulcer disease, gastritis, or a cardiac event given the radiation of pain to the left arm.} & {\hlgreen{Expanded differential diagnosis}} \\ \midrule
        
        {The doctor advised the patient to monitor their symptoms.} & {If symptoms worsen, \textbf{the patient should call the 15}.} & {\hlgreen{Explicit emergency guidance} \textit{(15 is the french 911)}} \\ \midrule

        {Home surveillance, and regular checks on the child's condition.} & {Home surveillance, frequent checks \textbf{every 3 hours}, and monitoring for any changes in the child's condition. } & {\hlgreen{clearer medical guidance}}  \\ \midrule

        {Stab wound to the upper arm.} & {Stab wound, \textbf{possible injury to underlying structures (e.g., nerves, blood vessels)}.} & {\hlgreen{Improved clinical specificity}} \\ \midrule

        {Triage Decision: Dispatch an ambulance.} & {Triage Decision: The patient is to be transported to the hospital by \textbf{car}.} & {\hlgreen{Adjusted transport method}} \\ \midrule

        {The patient's fourteen-year-old daughter is experiencing abnormal behavior, including difficulty speaking and walking, after a fight with her mother. \textbf{The daughter smells of alcohol.}} & {Altered mental status and \textbf{ataxia} in a 14-year-old female following a fight with her mother.} & {\hlpink{Lacks alcohol detail}, \hlgreen{Includes medical terminology}} \\ \midrule

        {The patient is the neighbor of the individual requiring medical attention. \textbf{The call was made from the neighbor's home}.} & {The caller is the patient's neighbor, \textbf{calling from the patient's home}.} & {\hlpink{Incorrect Call Context}}  \\ 
         
        \bottomrule    
    \end{tabular}
    }
    \caption{Further summary examples before and after fine-tuning on the \app selected summaries on the SimSAMU dataset. \textbf{Left}: Zero-Shot \app generates multiple summaries and selects the best ones. \textbf{Right}: Additional fine-tuning is applied to the selected summaries.}
    \label{tab:before_after_simsamu_long}
\end{table*}

\end{document}